\documentclass[a4paper,conference]{IEEEtran}

\usepackage{cite}
\usepackage{amsmath,amssymb,amsfonts}
\usepackage{algorithmic}
\usepackage{array}
\usepackage{graphicx}
\usepackage{url}
\usepackage{tabularx}
\usepackage{textcomp}
\usepackage{xcolor}
\def\BibTeX{{\rm B\kern-.05em{\sc i\kern-.025em b}\kern-.08em
		T\kern-.1667em\lower.7ex\hbox{E}\kern-.125emX}}

\begin{document}

\onecolumn
\textcopyright 2020 IEEE.  Personal use of this material is permitted.  Permission from IEEE must be obtained for all other uses, in any current or future media, including reprinting/republishing this material for advertising or promotional purposes, creating new collective works, for resale or redistribution to servers or lists, or reuse of any copyrighted component of this work in other works.
\twocolumn
\newpage

\title{Fully-Automated Packaging Structure Recognition in Logistics Environments}

\author{
	\IEEEauthorblockN{Laura D\"orr, Felix Brandt, Martin Pouls, Alexander Naumann}
	\IEEEauthorblockA{FZI Research Center for Information Technology\\
		Karlsruhe, Germany\\
		Email: \{doerr,brandt,pouls,naumann\}@fzi.de}
	}

\maketitle

\begin{abstract}
	Within a logistics supply chain, a large variety of transported goods need to be handled, recognized and checked at many different network points.
Often, huge manual effort is involved in recognizing or verifying packet identity or packaging structure, for instance to check the delivery for completeness.
We propose a method for complete automation of packaging structure recognition:
Based on a single image, one or multiple transport units are localized and, for each of these transport units, the characteristics, the total number and the arrangement of its packaging units is recognized.
Our algorithm is based on deep learning models, more precisely convolutional neural networks for instance segmentation in images, as well as computer vision methods and heuristic components.
We use a custom data set of realistic logistics images for training and evaluation of our method. 
We show that the solution is capable of correctly recognizing the packaging structure in approximately 85\% of our test cases, and even more (91\%) when focusing on most common package types.

\end{abstract}

\section{Introduction}
\label{sec:introduction}

\begin{figure}[htb]
	\centering
	\includegraphics[width=0.75\linewidth,trim={0 0.3cm 0 1.5cm},clip=true]{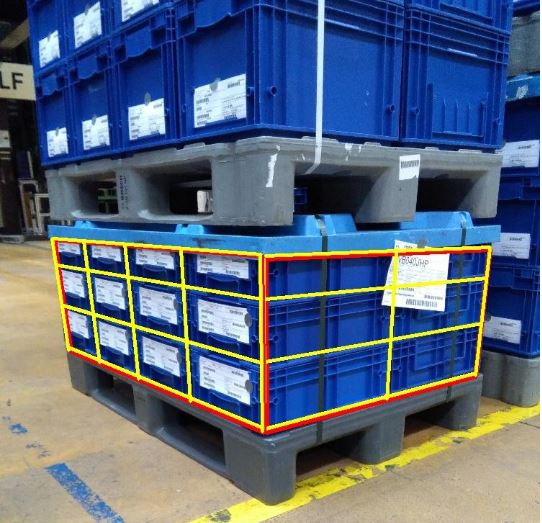}
	\caption{Illustration of the task of packaging structure recognition. The goal is to identify transportation units within single images and to recognize their package composition and structure. Red visualizes transport unit sides, yellow lines indicate packaging unit rows and columns.}
	\label{fig:psr_title}
\end{figure}

This work focuses on logistics packaging structure recognition. 
As this term is not commonly used, we define it as follows:
Packaging structure recognition is the task of visual recognition of a logistics transport unit and its essential building structure, i.e. the type, number and arrangement of unique packages.
This is illustrated in Fig. \ref{fig:psr_title}.

Packaging structure recognition is still a task frequently performed manually by humans in logistics supply chains (when counting rows and columns of packaging units and calculating the total number of packages on a transport unit).
We present fundamental work towards automatizing this process.
We develop a system which can, in a controlled environment like a factory or warehouse, independently perform packaging structure recognition, without any need for human interaction or guidance.
Basically, the system is a multi-step image processing pipeline operating on single RGB input images.
We make use of recent advances in computer vision methods for object detection and instance segmentation.
Namely, we use three convolutional neural networks (CNN) to detect delivery and packaging components.
We further apply traditional image processing techniques, and custom heuristics for information filtering and consolidation, to assemble a complex cognitive system for automated packaging structure recognition.
To emphasize the relevance of our work, we explain relevant logistics use-cases and possible benefits of our method.

Due to technical and organizational reasons, the proposed prototype system was trained and evaluated in a realistic restricted logistics environment.
In our evaluations, we show that the system's performance is feasible for application in such environments as the packaging structure of approximately 85\% of the test set's transport units are recognized correctly.
Still, the packaging components recognizable so far are strictly defined and modifications will require the employment of additional training data which may not be acquired easily.

The rest of the paper is organized as follows. 
We review related work in Section \ref{sec:related_work}.
In Section \ref{sec:problem}, we describe the problem and relevant use-cases, and we discuss potential benefits of our method as well as prerequisites and limitations.
Section \ref{sec:algorithm} contains detailed descriptions of the proposed algorithm, explaining all steps and components of our image processing pipeline.
The succeeding section \ref{sec:results} introduces the data used in training and evaluations.
Further, experiments and results for the whole pipeline and individual components are presented.
In section \ref{sec:summary}, we summarize our work and discuss future work, including possible improvements to our method.

\section{Related Work}
\label{sec:related_work}

Digitalization and process automation triggered by recent advances in technology are of increasing relevance, also in the sector of logistics and supply chain management, as recent publications show:
Wei et al. \cite{Wei19:DigitalTechnologiesLogistics} provide comprehensive insides into the ongoing digitalization of the logistics sector. 
The official positioning paper of the German Logistics Association (BVL) \cite{BVL:Digitalisierung} gives proof that this process is as well appreciated as it is inevitable.
Digitalization in logistics is very versatile regarding both applications and techniques.
Examples for applications are automated guided vehicles (AGV), cross-company communications and information systems, predictive maintenance and manufacturing robotics.
Technological foundations of modern applications range from barcode recognition and data digitization methods by various sensors, over optimization and data analytics to modern methods like block chain, deep learning and virtual reality.
Our contribution utilizes means of image processing and artificial intelligence. 
Borstell \cite{Borstell18:ImageProcessingLogistics} gives an overview of how these methods have recently been used in various logistics applications.

The idea of automating parcel recognition is not new to the logistics sector as various companies work on related applications.
A machine vision system by Zetes \cite{Solution:zetes} offers automated reading of barcodes and labels on logistics units and archiving of transport unit images.
Vitronic \cite{Solution:vitronic} offer a very similar machine vision systems for the automation of goods receipt and shipping. 
Package volumes can also be measured by additional engagement of laser scanners.
Another camera-based system for automated barcode reading, sorting and dimensioning is offered by cognex \cite{Solution:cognex}.
Logivations \cite{Solution:Logivations} offer a vision system which additionally counts and measures objects within a logistics units and can be trained by the user to recognize custom objects based on their appearance.
None of these systems aims to capture the total number of packages in an assembled transport unit, as opposed to our work.
The number of scientific publications regarding logistics automation use-cases are very low.
To our knowledge, no other scientific publication focusing on packaging structure recognition exists.
Fraunhofer IML have developed a system for image-based automated counting of carriers \cite{Solution:Fraunhofer_Ladungstraeger}, but do not provide further information on the technologies involved.

Technically, our work makes use of recent advances in the image processing problems of object detection and instance segmentation.
We employ Mask-RCNN, a state-of-the-art neural network for instance segmentation, by He et al. \cite{He2017:MaskRCNN} for segmentation of logistics components.
As feature extraction layers within the Mask R-CNN segmentation network, we use Inception-v2 \cite{Ioffe15:BatchNorm}.
An overview over deep learning based instance segmentation, as well as further introductions to artificial neural networks, is given by Minaee et al. \cite{Minaee2020:ImageSegmentationSurvey}.

As the amount of data required to train a image instance segmentation model from scratch is tremendous, a remedy often applied is transfer learning.
This term refers to the idea of pre-training a prediction model on a large dataset of a general task, before fine-tuning the previously learned weights on rather few use-case specific data.
Not at all a particular deep learning or image processing technique, transfer learning is used in various contexts \cite{pan2009:transferlearningsurvey}.
The method has been applied to image object detection or instance segmentation tasks and use-cases \cite{azizpour2015:transferlearning} from various domains, such as medical imaging \cite{shin2016:transferMedicalImaging}, airport security \cite{akccay2016:transferAirportSecurity} and environmental engineering \cite{gao2018:transferEnviornment}, to name only a few examples.
For most image processing applications, pre-training is performed either on the image classification dataset Image-Net \cite{russakovsky2015:imagenet}, \cite{huh2016:transferlearningImagenet}
or on the COCO dataset for object detection tasks \cite{Coco2018}.
The latter is also the case in our work.

\section{Problem and Prerequisites}
\label{sec:problem}

In this section, we give insides into the logistics transportation setting the problem and solution are drawn from.
First of all, we define important terms which are frequently used throughout this paper.
The logistics use-case in which this work was created is described in detail and, additionally, other relevant use-cases are mentioned.
The section concludes with prerequisites and limitations to our image processing pipeline.

\subsection{Terms and Definitions}

\subsubsection{Packaging Unit}
\begin{figure}
	\centering
	\includegraphics[width=0.4\linewidth,trim={0 1.5cm 0 1.5cm},clip=true]{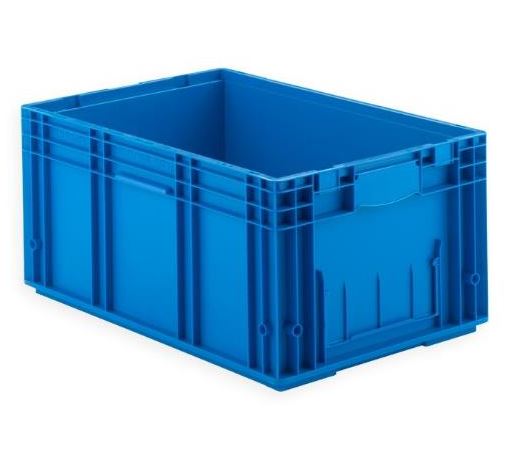}
	\caption{Example of a KLT packaging unit \cite{VDA:KLT}.}
	\label{fig:KLT}
\end{figure}

Packaging units are used to hold one or multiple items allowing for standardized goods transportation. 
A large variety of containers is used for different kinds of transport goods depending on their size, weight, material, number or simply on the industrial sector.
In our case, we focus on a small subset of highly standardized packaging units, which is defined by the German Association of the Automotive Industry (VDA): the small load carrier system (KLT) \cite{VDA:KLT}.
An example for a KLT packaging unit is shown in Fig. \ref{fig:KLT}.

\subsubsection{Base Pallet}
To allow for standardized handling of logistics transport units, independent from the type of packaging unit used, logistics goods are shipped on standardized base pallets. 
The most prominent example thereof is the EUR-pallet (or EPAL-pallet) \cite{EPAL:pallet}.

\subsubsection{Logistics Transport Unit}
\begin{figure}
	\centering
	\includegraphics[width=0.3\linewidth,trim={0 8cm 0 5.5cm},clip=true]{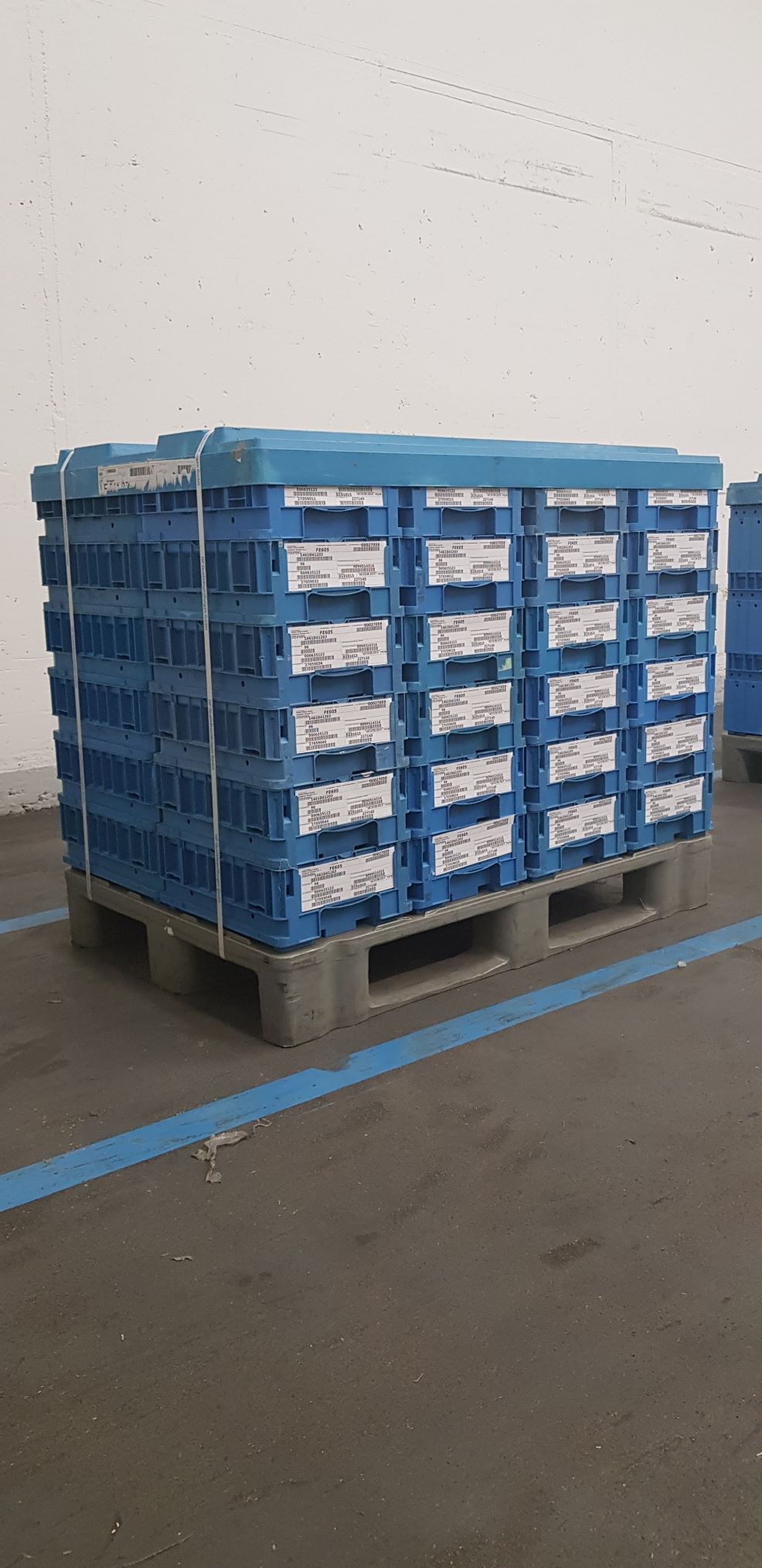}
	\includegraphics[width=0.3\linewidth,trim={0 8cm 0 5.5cm},clip=true]{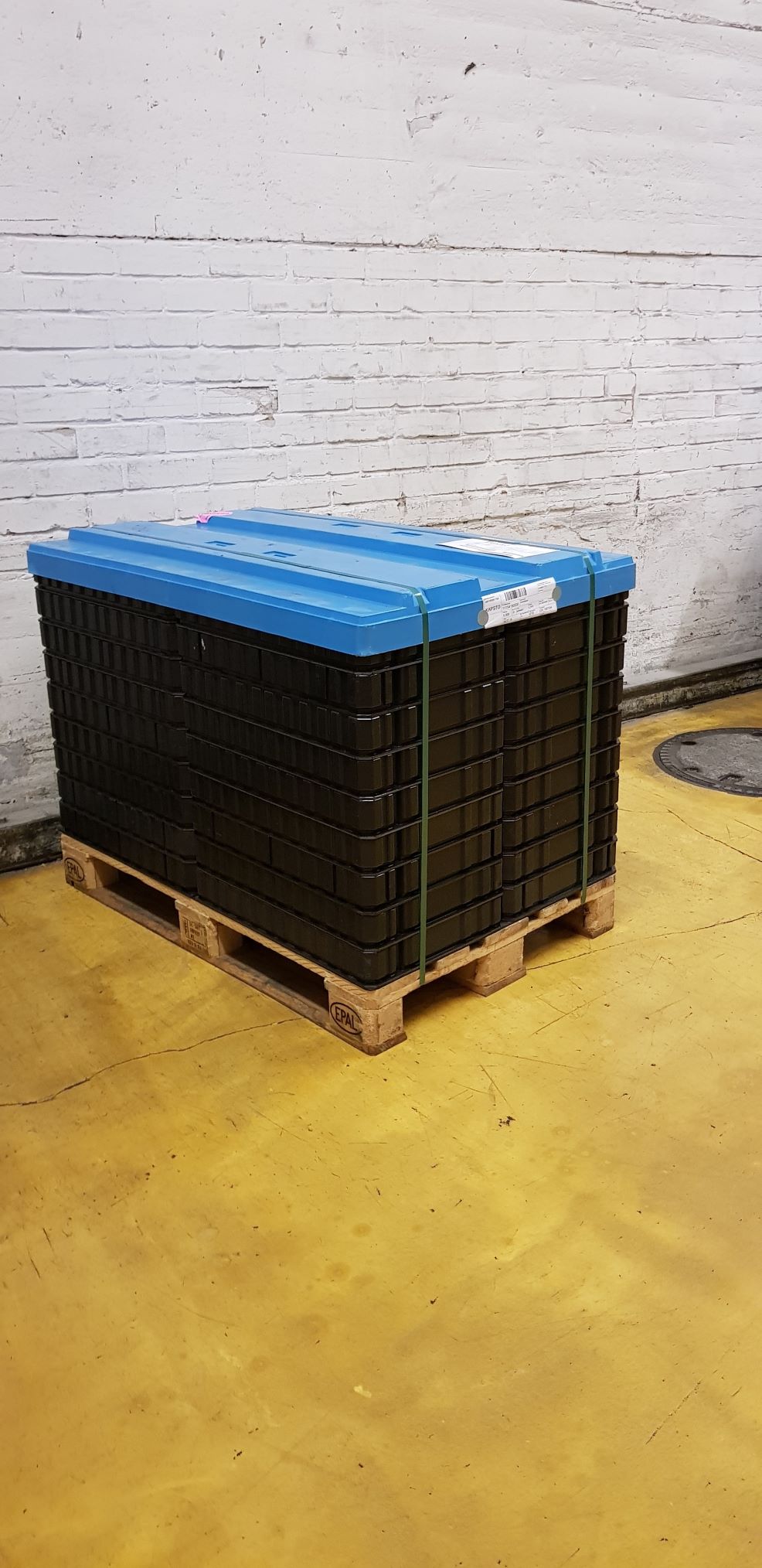}
	\includegraphics[width=0.3\linewidth,trim={0 8cm 0 10cm},clip=true]{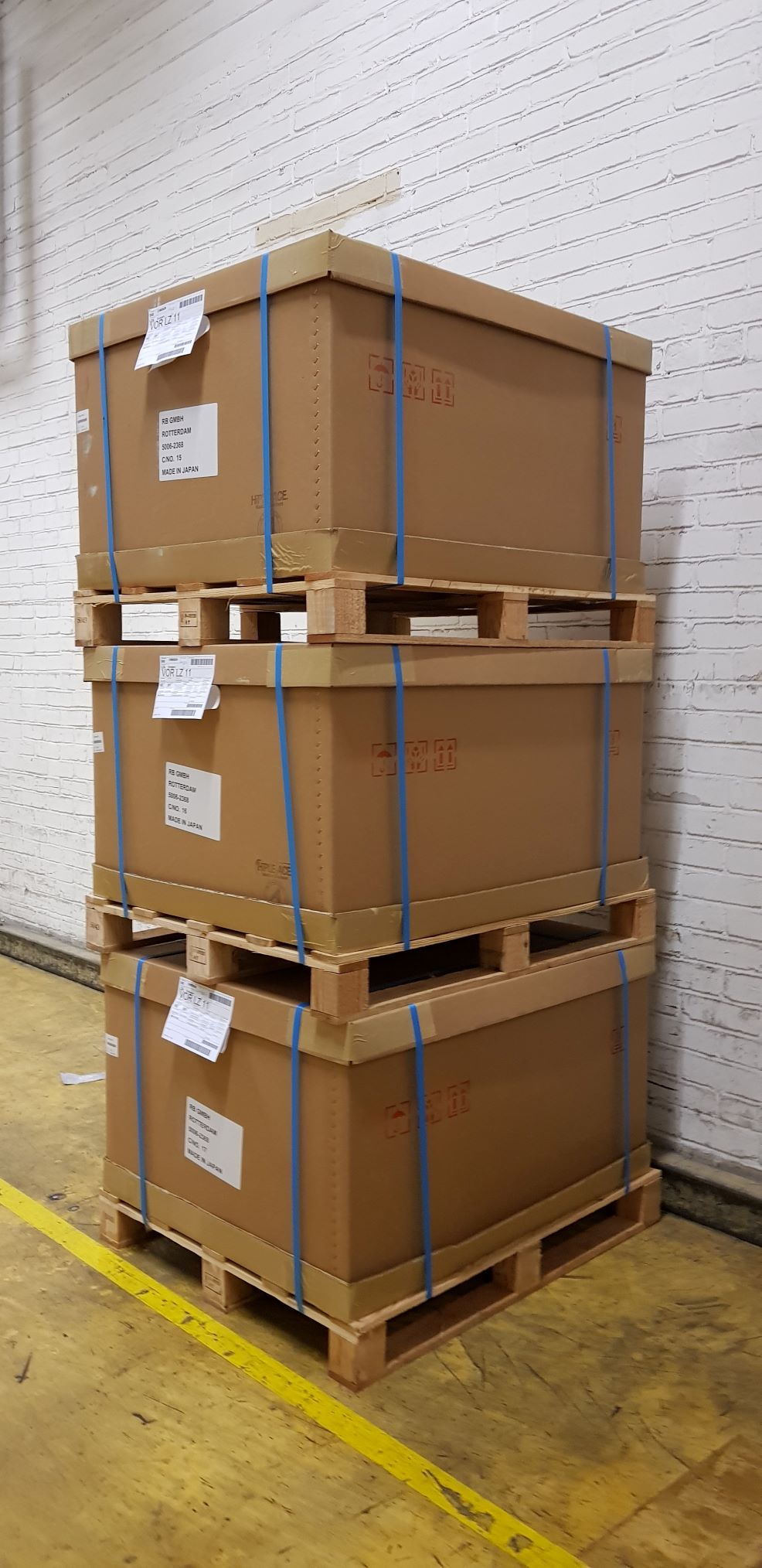} \\
	(a) \hspace{0.25\linewidth} (b) \hspace{0.25\linewidth} (c) \\
	\caption{Examples for logistics transport units with different packaging components.}
	\label{fig:transportunits}
\end{figure}

When speaking of a logistics transport unit, we address a fully-packed, labeled and shipping-ready assortment of goods. 
Such units are usually composed of a base pallet, a set of logistics packaging units and a pallet lid. 
The appearance, size and material of packaging units can vary largely.
Often, optional components like transparent foils, security straps and transport labels can also be attached to transport units.
Some example images for logistics transport units are shown in Fig. \ref{fig:transportunits}.

\subsection{Motivation and Use-Case Descriptions}
In the following we aim to motivate and demonstrate the need for and benefits of automated packaging structure recognition.
Therefore, we exemplary name a few relevant logistics use-cases to demonstrate re-usability of our method:
\begin{enumerate}
	\item Automated incoming goods transaction
	\item Automated outgoing goods control
	\item Automated empty packaging counting
\end{enumerate}

We focus on the first of these use-cases as our algorithm was primarily developed for such a setting.
The other use-cases are only briefly sketched to argue our method's relevance.

\subsubsection{Automated incoming goods transaction}
Incoming goods processes are an essential part of every logistics supply chain, which are, for instance, described in \cite{Furmans2019:Wareneingang}.
On receipt of logistics goods, the execution of corresponding booking operations is mandatory.
In Germany, it is enforced by law (German Commercial Code (HGB) §377) that incoming goods have to be, at least externally, checked for completeness and damages immediately after reception.
Commonly, the goods booking task is preceded by manual inspection of the transport units received: 
Transport labels are read, either manually or using a scanning device, packages are counted and the unit is checked for damages and tampering.
In some cases, packages are opened for more thorough examination.
The steps included in incoming goods checks can vary.
Nonetheless, package number verification is necessary in all applicable cases. 

While systems for the automated detection and reading of visible transport labels and barcodes on transport units exist, 
the whole process can not be covered by such systems:
In most cases, not all packages and transport labels can be captured in a single image as they may be attached to different sides of the transport unit or may be occluded.
If packages are arranged in a 3D pattern, it is not possible to picture all packages at once and counting does not suffice to find the total number of packaging units.
Our method tackles this problem by recognizing the transport unit's distinct sides and counting packages individually for each side.
As units are not only counted, but also their arrangement is captured, the total number of packages can be calculated.

\subsubsection{Automated outgoing goods control}
Similar to incoming goods controls, outgoing goods need to be checked similarly before dispatching.
On the one hand, individual transport units are checked for completeness, integrity and packaging instruction compliance.
On the other hand, it has to be ensured that the transport units take the right track and are loaded onto the designated means of transport.


\subsubsection{Automated empty packaging counting}
As KLTs are part of a deposit system, such empty transport units are still valuable resources which need to be organized and repurposed:
Once emptied, transport units are often cleaned and, subsequently, relocated for further use or temporarily stored.
Hereby, they can be assembled to transport units or handled individually.
In such cases, our method can be used to enable automatized logging, booking or stocktaking of KLT empties.

\subsection{Problem Formulation}
We define the basic task of packaging structure recognition as the challenge of inferring a logistics transport unit's packaging structure from a single image of that unit.
Here, the packaging structure consists of the following information:
\begin{itemize}
	\item Type and number of packaging units
	\item Arrangement of packaging units
	\item Type of base pallet
\end{itemize}

In this work, the types of packaging units and base pallets distinguishable are limited to a small set relevant in the context of our test data. 
In many logistics use-cases, the limitation to known packaging units and components does not compromise the applicability of our algorithm as the vast majority of transport units comply to explicit packaging standards and unknown packaging components are not to be expected.

Multiple extensions and refinements to our method are possible and intended, but not in the scope of this work.
Some of these extensions are necessary to fully cover the use-cases described in the previous subsection.
For example, the additional detection of packaging components such as lids, security straps or transparent foils, is intended to allow for automated checks of packaging instruction compliance.

\subsection{Prerequisites and Limitations}
\label{sec:prerequisites}

The proposed prototype system was trained to work in a restricted logistics environment.
All tests and evaluation were performed in the same setting.
Thereby, restricted means that all types of transport units and components, as for instance packages and base pallets, are known before-hand and special requirements regarding the input image exist.
The reasons for these restrictions are, on the organizational side, the difficulties in acquiring realistic, annotated training and test data of variant logistics environments.
On the technical side, difficulties arise as due to the application of learning methods which can only generalize to what they have seen in training.
Image instance segmentation learning models recognize and distinguish only between those classes previously seen in training data.
Evidently, there is research on few-shot or zero-shot learning working towards training algorithms which aim to distinguish instances of new classes, after having seen only very few or even no examples of that class in training \cite{Wang2019:fewshotlearning}.
Still, in the scope of this work, we stick to well-proven deep learning approaches for object detection, which require all the object classes to be known beforehand.
All perquisites and limitations assumed valid are summarized in the following.

\subsubsection{Material Restrictions}
In logistics supply chains, the package types used can vary largely, depending on the industry sector, the transported goods and the companies involved.
We limit our models for package recognition to a well-defined subset of package types, in accordance with our data set.
The package types present in these images are standardized euro transport packages (KLT) of different sizes and colors (see Fig. \ref{fig:transportunits} (a)) and so-called tray packages (see Fig. \ref{fig:transportunits} (b)).

Similar to package types, also the base pallet types present in the data have to be known beforehand.
Our data contains two different types of base pallets: wooden EPAL Euro pallets and plastics reusable pallets of the same size (1200 x 800 mm).

\subsubsection{Packaging Restrictions}
We focus on uniformly packed transport units.
This means, each transport unit may be composed of only one single package type and packages are ordered regularly in full levels. 
There are no gaps between neighboring package units and each level of package units has the same number of packages.
In this case, the complete packaging structure can be inferred by only observing one image of the transport unit, if taken from the right perspective.
For non-uniformly packed transport units, the task is in general not solvable as the information contained in one image is not sufficient to infer the unit's packaging structure.

\subsubsection{Imaging Restrictions}
In order for one or more transport units to be correctly recognized in an image, the image needs to meet the following conditions:
\begin{itemize}
	\item \textbf{Orientation}: Transport units need to be upright in the image. Real-world vertical lines should be roughly parallel to the vertical image boundaries.
	\item \textbf{Perspective}: Transport units are not shown in a frontal perspective, but in such a way that two sides are clearly visible and completely covered by the image. 
	\item \textbf{No occlusions}: Relevant transport units are completely visible within the image. No parts of the unit are occluded by other objects or lie outside the image boundaries.
\end{itemize}

Note that, if both the orientation and perspective criteria are met, a left-most and a right-most transport unit side can be clearly identified for each transport unit.

\section{Algorithm and Implementation}
\label{sec:algorithm}

The algorithm extracting the packaging structure of a transport unit from a single image is described in this section.
The algorithm consists of three essential steps: 

\begin{enumerate}
	\item Transport Unit Detection: Inter-unit Segmentation
	\item Packaging Unit Detection: Intra-unit Segmentation
	\item Result Consolidation and Refinement
\end{enumerate}

\begin{figure*}
	\begin{center}
		\includegraphics[width=0.23\linewidth,trim={0 4.1cm 0 4cm},clip=true]{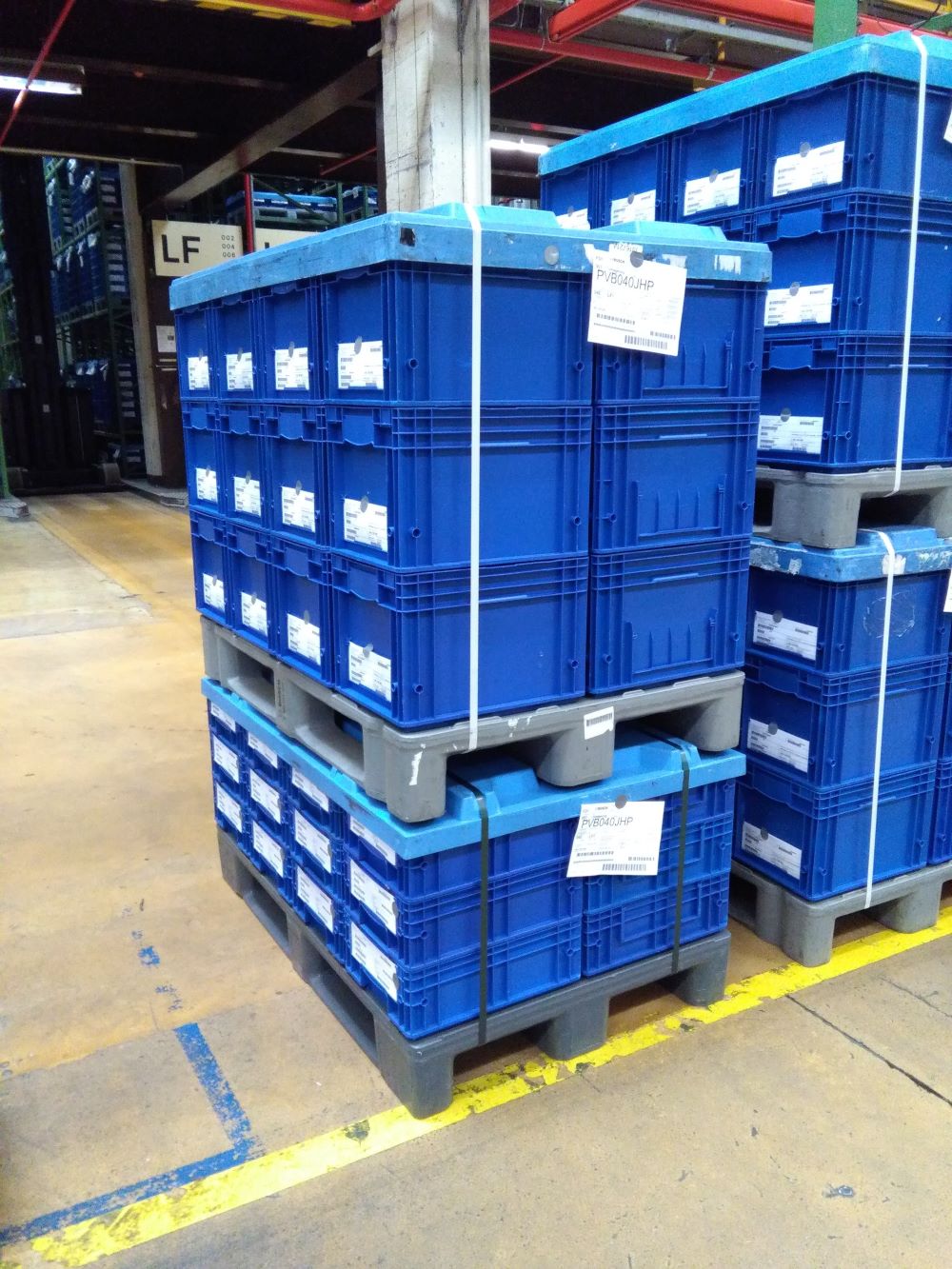}
		\includegraphics[width=0.23\linewidth,trim={0 4.1cm 0 4cm},clip=true]{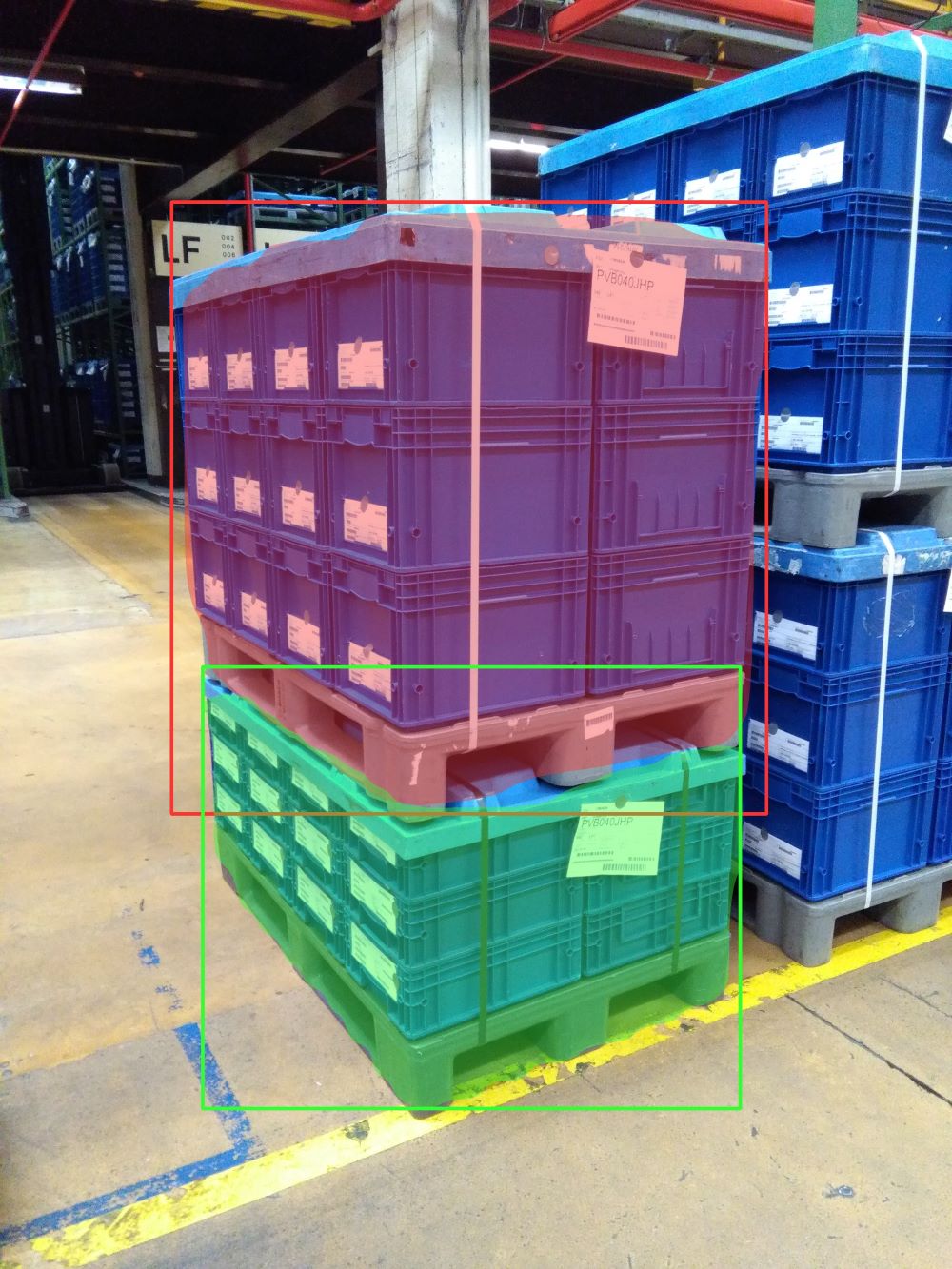}
		\includegraphics[width=0.23\linewidth]{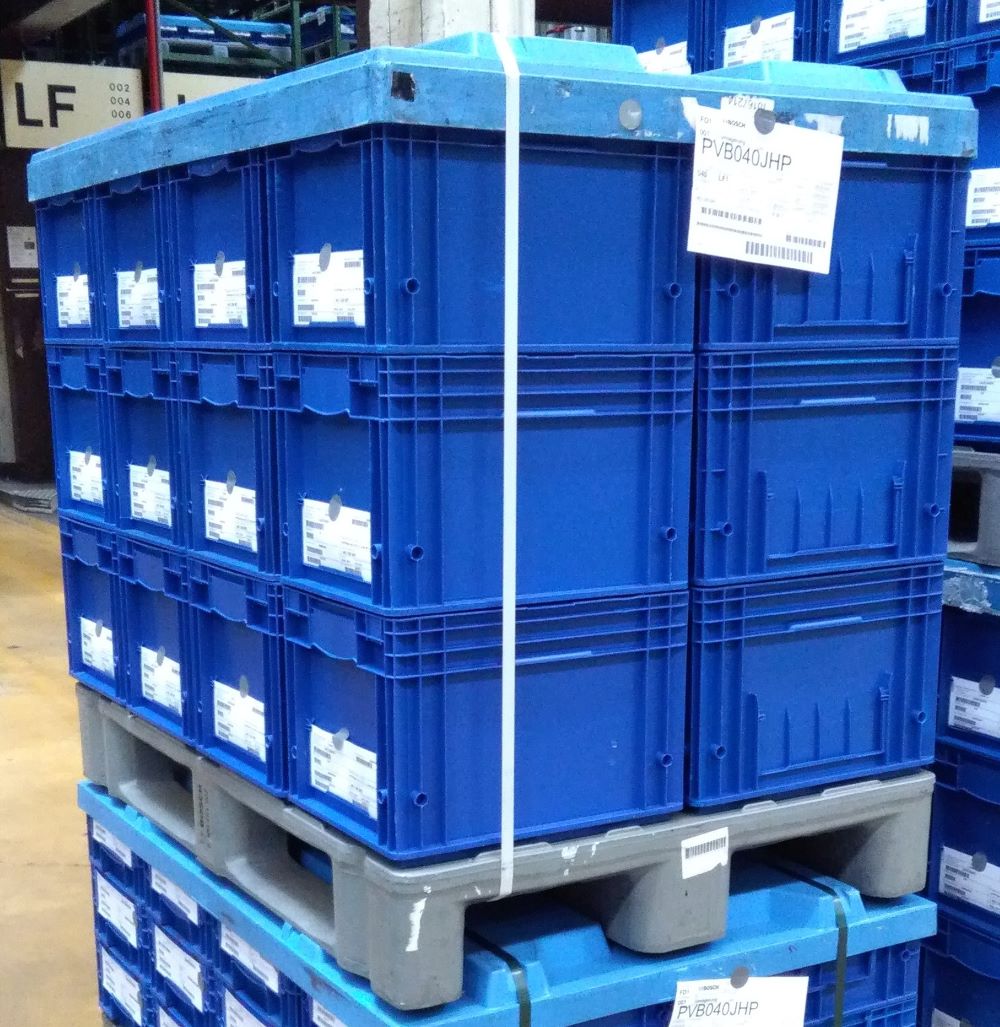}
		\includegraphics[width=0.23\linewidth]{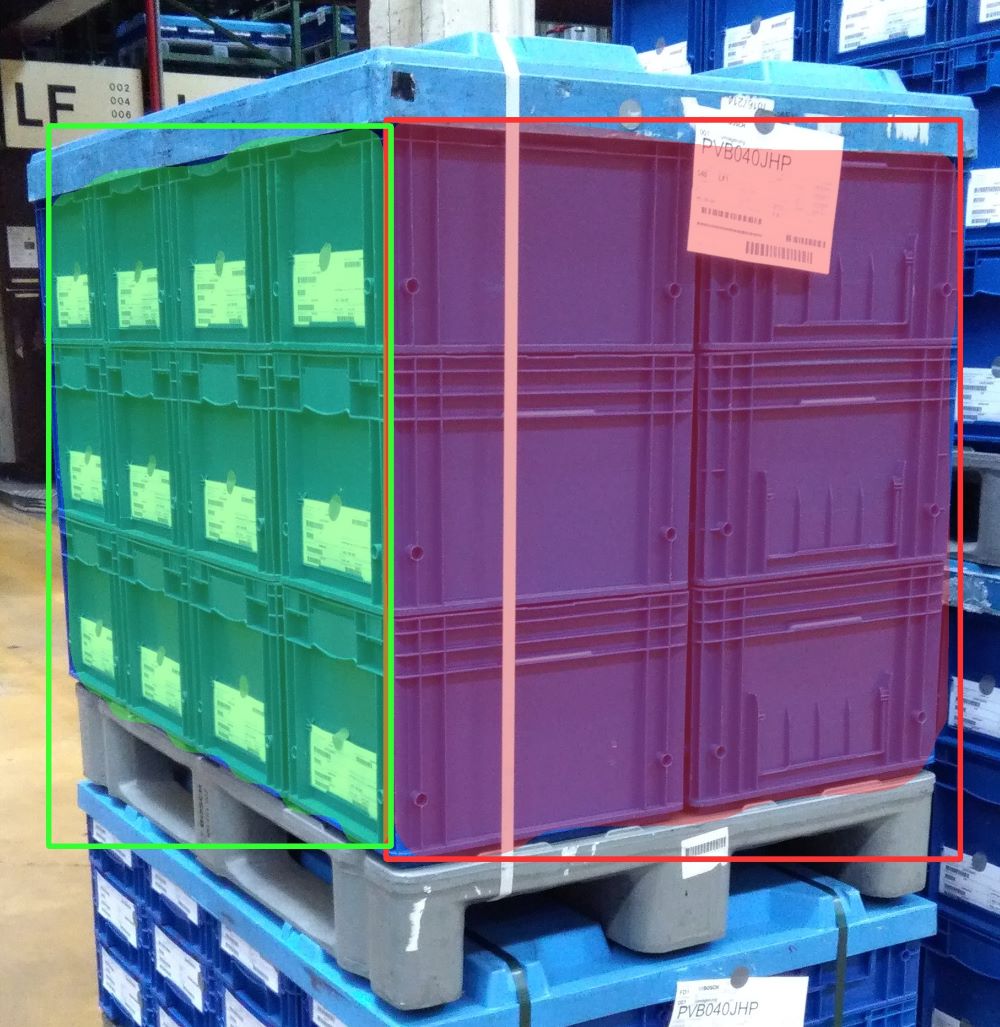} \\
		\mbox{(a)} \hspace{0.2\linewidth} \mbox{(b)} \hspace{0.2\linewidth} \mbox{(c)} \hspace{0.2\linewidth} \mbox{(d)} \\
		\includegraphics[width=0.23\linewidth]{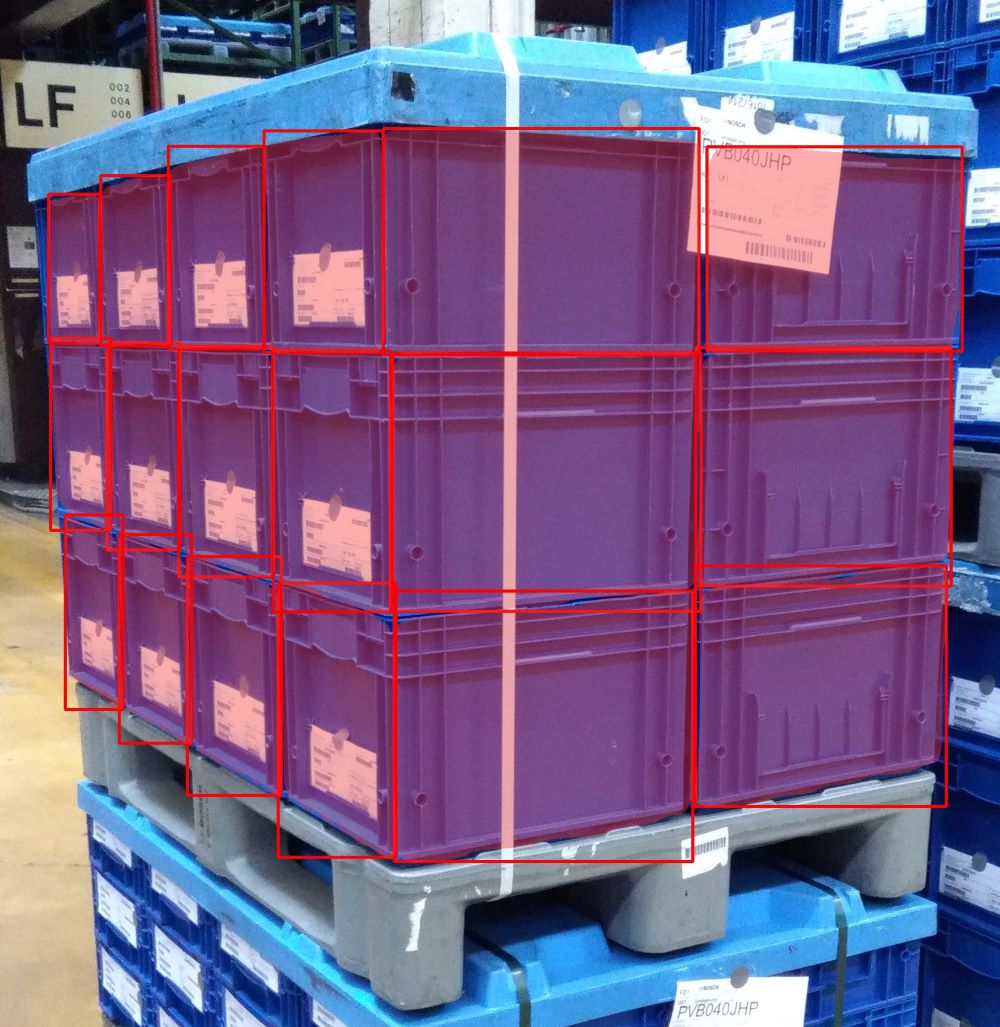}
		\includegraphics[width=0.23\linewidth]{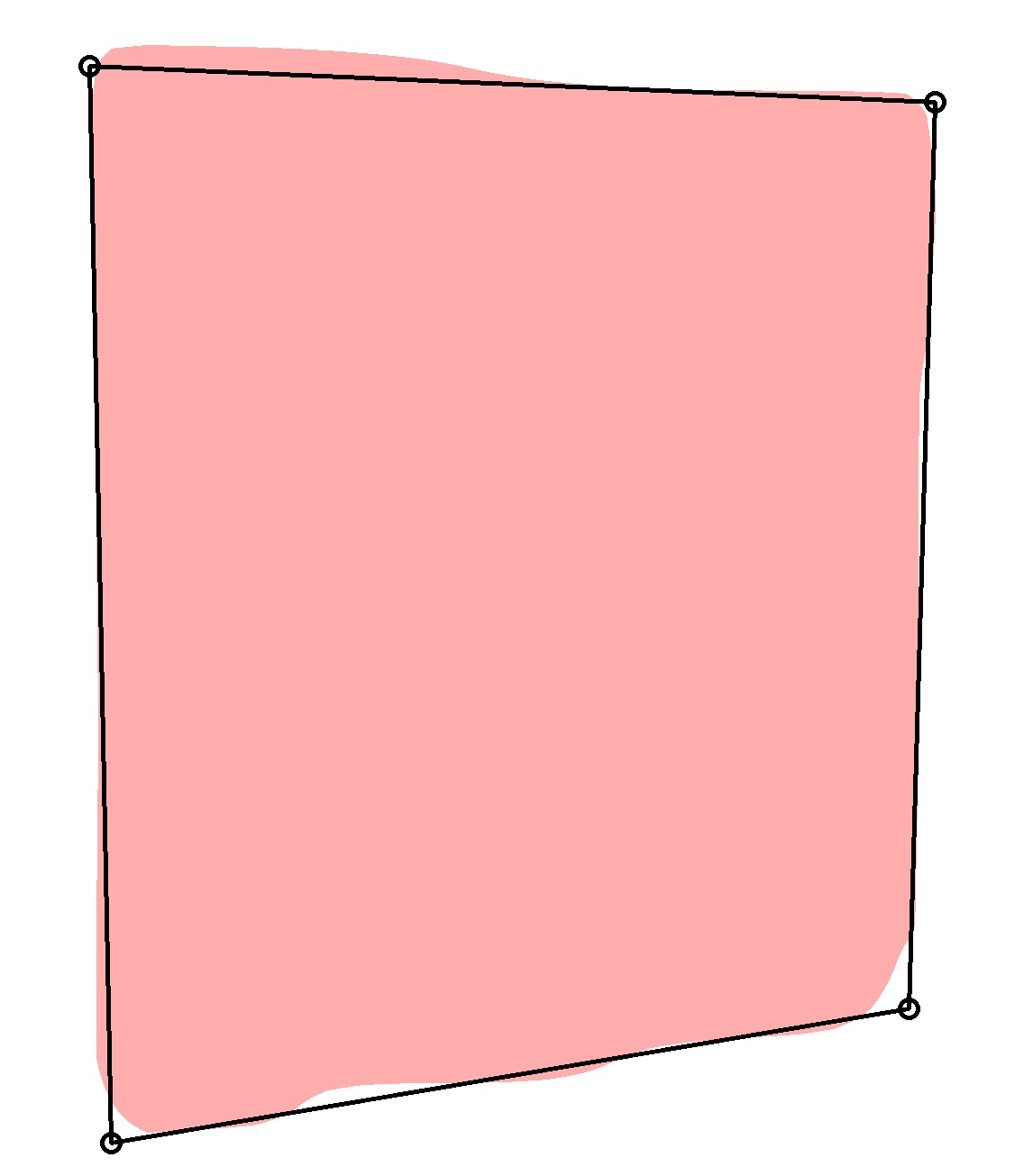}
		\includegraphics[width=0.23\linewidth]{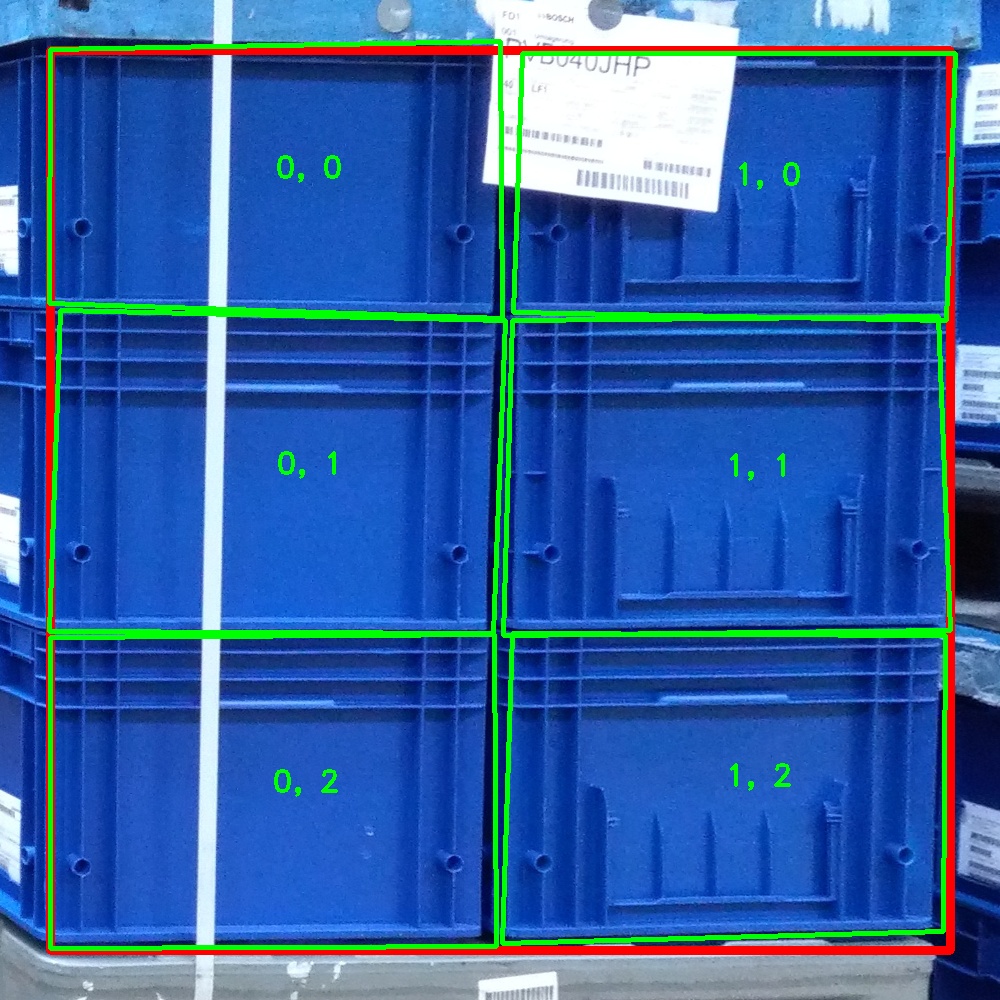}
		\includegraphics[width=0.23\linewidth]{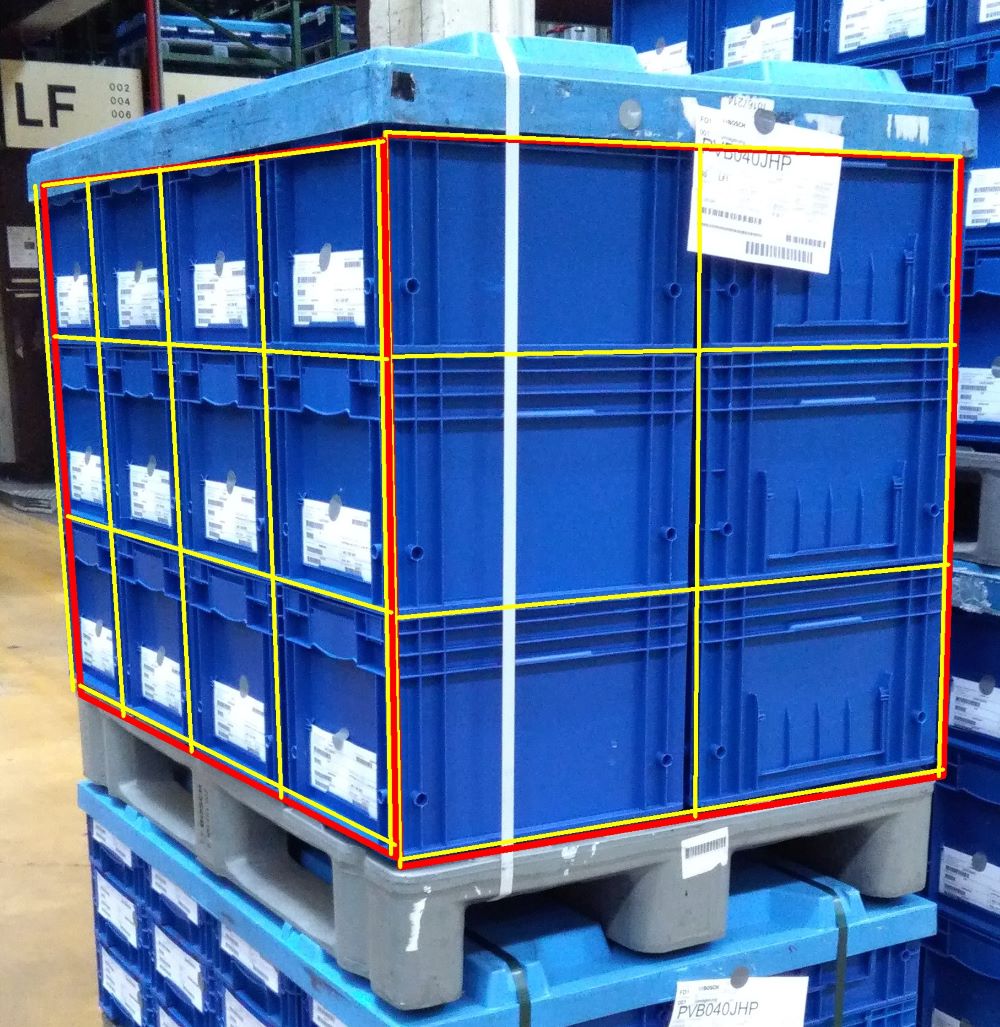} \\
		\mbox{(e)} \hspace{0.2\linewidth} \mbox{(f)} \hspace{0.2\linewidth} \mbox{(g)} \hspace{0.2\linewidth} \mbox{(h)} \\
		\caption{Algorithm illustration. 
			(a) Input image. 
			(b) Inter-unit segmentation: Detected transport units with masks. 
			(c) Cropped image for the top-most transport unit. 
			(d) Intra-unit segmentation: Detected transport unit sides with masks. 
			(e) Intra-unit segmentation: Detected packaging units with masks.
			(f) Right transport unit side mask and approximated tetragonal shapes.
			(g) Right rectified transport unit side with rectified and 2d-indexed package unit regions.
			(h) Visualization of the recognized packaging structure.}
		\label{fig:algorithm_illustration}
	\end{center}
\end{figure*}

First of all, an \textbf{inter-unit segmentation} is performed to detect relevant transport units within the input image. 
A convolutional neural network (CNN) is used to find and segment all transport units which are completely visible in the image.
In the next step, the \textbf{intra-unit segmentation} is applied to each cropped transport unit image. 
Here, two CNNs are used to find the unit's pallet and to detect the two distinct sides as well as all packaging units.
Computer vision methods are used to \textbf{consolidate and refine} the results.
The whole process is illustrated by means of a single test image in Fig. \ref{fig:algorithm_illustration}.

\subsection{Inter-Unit Segmentation}
Given an input image as described above, the first step in our processing pipeline is the detection and segmentation of all relevant transport units within the image.
Precisely speaking, we want to extract how many transport units are fully contained in the image and which image regions and pixels belong to which of those transport units.
To solve this task, a deep learning instance segmentation model is used.
Simple sanity checks are performed on the instance segmentation model's predictions:
Apart from a confidence thresholding, the detections are checked for a minimum size within the image (in both directions independently) and detections, which have high bounding box overlaps with other higher-confidence detections, are dropped.
Fig. \ref{fig:algorithm_illustration} (b) shows exemplary results for the transport unit segmentation.
For each detected transport unit, a corresponding image crop is created. See Fig. \ref{fig:algorithm_illustration} (c).

\subsection{Intra-Unit Segmentation}
For each image crop output by the inter-unit segmentation, an intra-unit segmentation is performed independently.
Intra-unit segmentation aims to further segment the transport unit image. The objective is to identify the following regions:
\begin{itemize}
	\item Base pallet
	\item Exactly two transport unit sides
	\item Package unit faces (not complete units)
\end{itemize}
Apart from region information, the types of base pallet and packages is also determined.
One very important aspect here is the identification and differentiation of the two visible, orthogonal transport unit sides.
Only if both sides are segmented correctly and the detected package unit faces are assigned to the correct transport unit sides, the total package unit number can be calculated accurately.
Again, simple consistency checks are performed:
All detection regions are checked for overlaps with the previously extracted transport unit region:
Only detections with a bounding box of which at least 60\% lies within the transport unit's bounding box are kept.
Further, detections of low confidence or transport unit sides with inadequate region sizes are dropped.
For each transport unit, it is ensured that exactly one base pallet and two transport unit sides are found.
Otherwise, the algorithm aborts without returning any result.
Transport unit side segmentation and packaging unit segmentation are illustrated in Fig. \ref{fig:algorithm_illustration} (d) and (e), respectively.

\subsection{Information Consolidation}

The goal of this final step is to use the previously extracted segmentation information to determine the transport unit's packaging structures, i.e. to calculate the total number of packages.
The algorithm works regardless of the type of packaging units present.
The following steps are performed:

\begin{enumerate}
	\item Assign each package unit to one transport unit side
	\item Refine region segmentation
	\item Calculate package number
\end{enumerate}

Each of the above steps is explained in the following.

\subsubsection{Assign each package unit to one transport unit side}
For each packaging unit, the intersection of its mask with both transport unit sides masks is computed. 
If at least one of these intersections is not empty, it is assigned to the side with larger absolute mask intersection size.
Packaging units not intersecting any of the transport unit sides are dropped.

\subsubsection{Refine region segmentation}
Package unit and transport unit side segmentation masks are cut off outside the transport unit's mask as the transport unit segmentation is assumed to be more robust and accurate than the intra-unit segmentation.
This assumption is supported by the segmentation model evaluations deduced in Section \ref{sec:results}.
Further, transport unit side masks are extended by the union of all packaging unit masks assigned to this side, which has empirically shown to be beneficial for the overall results.

\subsubsection{Calculate package number}
To calculate each transport unit side's package number in vertical and horizontal direction, the average packaging unit size is computed.
Due to perspective distortions, transport unit sizes of identical packages vary essentially within the image depending on their position relative to the camera.
Packaging unit and transport unit side regions are rectified to overcome this issue:
To perform rectification, the mask of each transport unit side is approximated by a tetragon shape.
To approximate the polygon describing the side's mask, an optimization problem minimizing the region difference for the detected transport unit side mask and the shape described by four corner points is considered:
\begin{equation}
	\min_{x_{1}, x_{2}, x_{3}, x_{4} \in P}{\sum_{(i, j) \in P} {\left| s_{(i,j)} - t(x_1, x_2, x_3, x_4, (i, j)) \right|} }
\end{equation}
where $P = \{0,...,m\} \times \{0,...,n\}$ is the input image's pixel space,
$s \in \{0, 1\}^{m \times n}$ is the binary mask describing the transport unit side as output by the segmentation model
and $t: P^5 \rightarrow \{0, 1\}$ with $t(x_1, x_2, x_3, x_4, (i, j)) = 1$ if pixel $(i,j)$ lies within the tetragon described by the corner points $(x_1, x_2, x_3, x_4)$.
The optimization problem is solved numerically, using OpenCV's implementation of the Douglas-Peucker algorithm \cite{Douglas1973:AlgorithmsFT} on the polygon described by mask $m$ to find suitable starting points.
For a single example image, the whole process and the results are illustrated in Fig. \ref{fig:algorithm_illustration} (f).

Using the result as approximation for the transport unit side's four corner points, the detections associated with this transport unit side are remapped in such a way that the transport unit side is described by a rectangle of size ${s}_{v} \times {s}_{h}$. This is illustrated in Fig. \ref{fig:algorithm_illustration} (g).
The height ${ps}^i_h$ and width ${ps}^i_v$ of each packaging unit $i$ is approximated as the corresponding bounding box' size after rectification.
Now, the average packaging unit height $\tilde{{ps}_v}$ and width $\tilde{{ps}_h}$, relative to the transport unit side's size, can be calculated.
The horizontal and vertical package numbers $n_{h}$ and $n_{v}$ are computed as:
\begin{eqnarray}
	n_h = \lfloor \frac{s_h}{\tilde{{ps}_h}} + 0.5 + \delta_1 \rfloor \\
	n_v = \lfloor \frac{s_v}{\tilde{{ps}_v}} + 0.5 + \delta_2 \rfloor
\end{eqnarray}

Hereby, the additional summands $\delta_1, \delta_2 > 0$ account for the empirically discovered tendency towards over-estimation of package unit sizes.
In our experiments, the values were set to $\delta_1 = 0.05$ and $\delta_2 = 0.15$. 
In the computation of the vertical package number the additional summand $\delta_2$ is larger than $\delta_1$, which can be explained as follows:
The packaging units in a transport unit's top row are frequently partly occluded by pallet covers (for example, see Fig. \ref{fig:algorithm_illustration}).
Thus, these packaging units are not completely visible and their regions detected by the segmentation model are smaller than those of the other rows of packaging units.
The larger choice of $\delta_2$ helps to account for these size underestimations.
Once the horizontal and vertical package numbers are calculated for both transport unit side, the overall package number can be determined.
If the vertical package numbers for the two side do not coincide, the algorithm stops without returning any package number result.
Otherwise, the package number is calculated as
\begin{equation}
	n = n_h^{(l)} \cdot n_h^{(r)} \cdot n_v
\end{equation}
where $n_h^{(l)}$ and $n_h^{(r)}$ are the horizontal package number for the left and right transport unit side within the image and $n_v = n_v^{(l)} = n_v^{(r)}$ is the vertical package number.

Fig. \ref{fig:algorithm_illustration} (h) exemplary shows a comprehensive illustration of the packaging structure recognition results.

\section{Experiments and Results}
\label{sec:results}

\begin{figure*}[htb]
	\centering
	\includegraphics[width=0.19\linewidth,trim={0 4cm 0 4cm},clip=true]{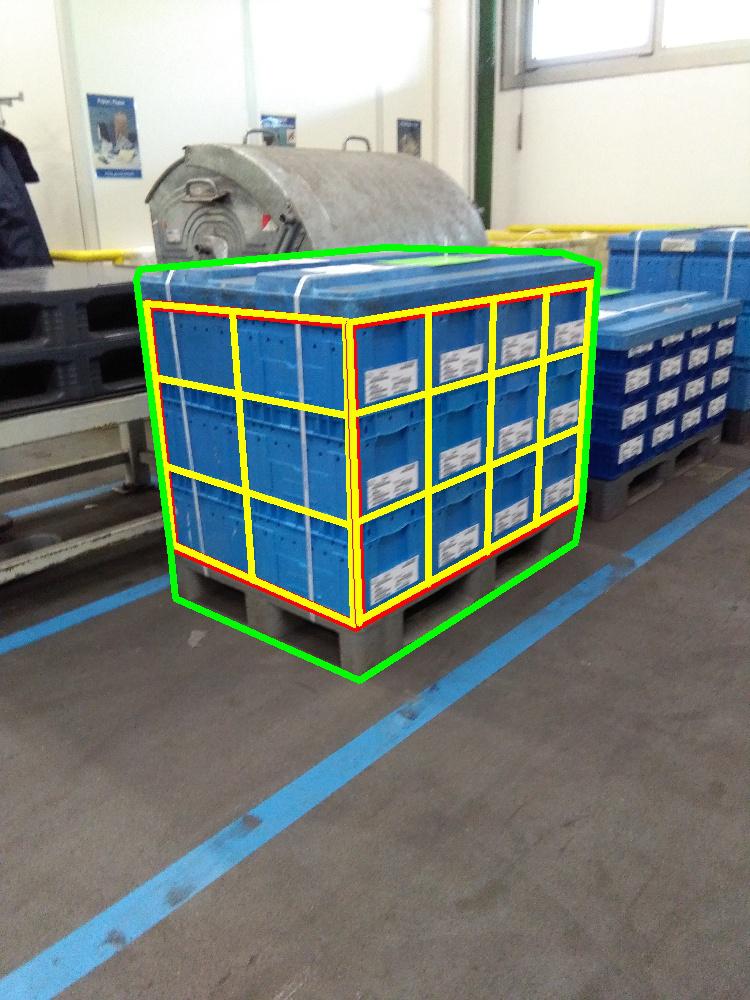}
	\includegraphics[width=0.19\linewidth,trim={0 4cm 0 4cm},clip=true]{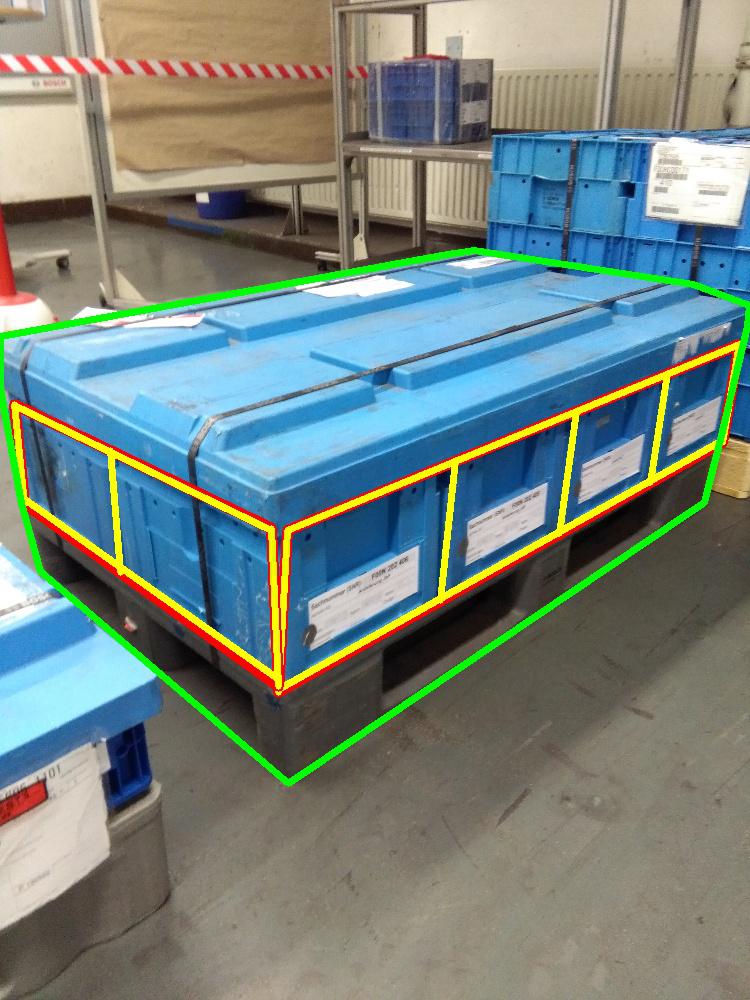}
	\includegraphics[width=0.19\linewidth,trim={0 2.5cm 0 2cm},clip=true]{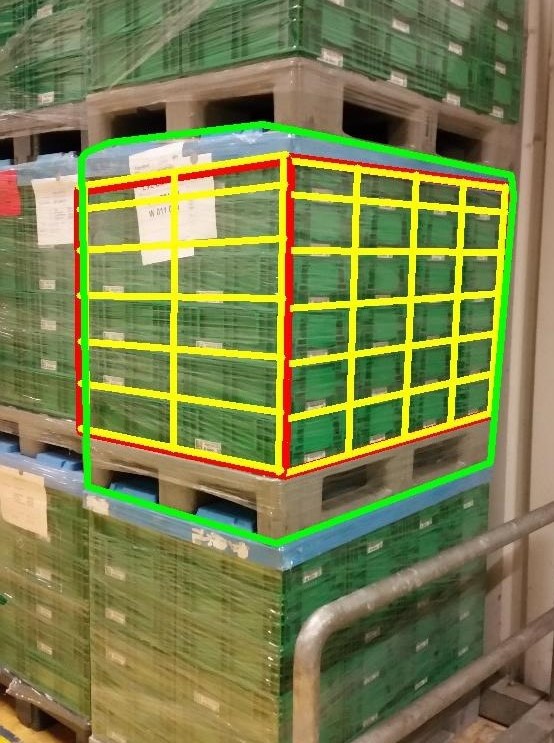}
	\includegraphics[width=0.19\linewidth,trim={0 4cm 0 4cm},clip=true]{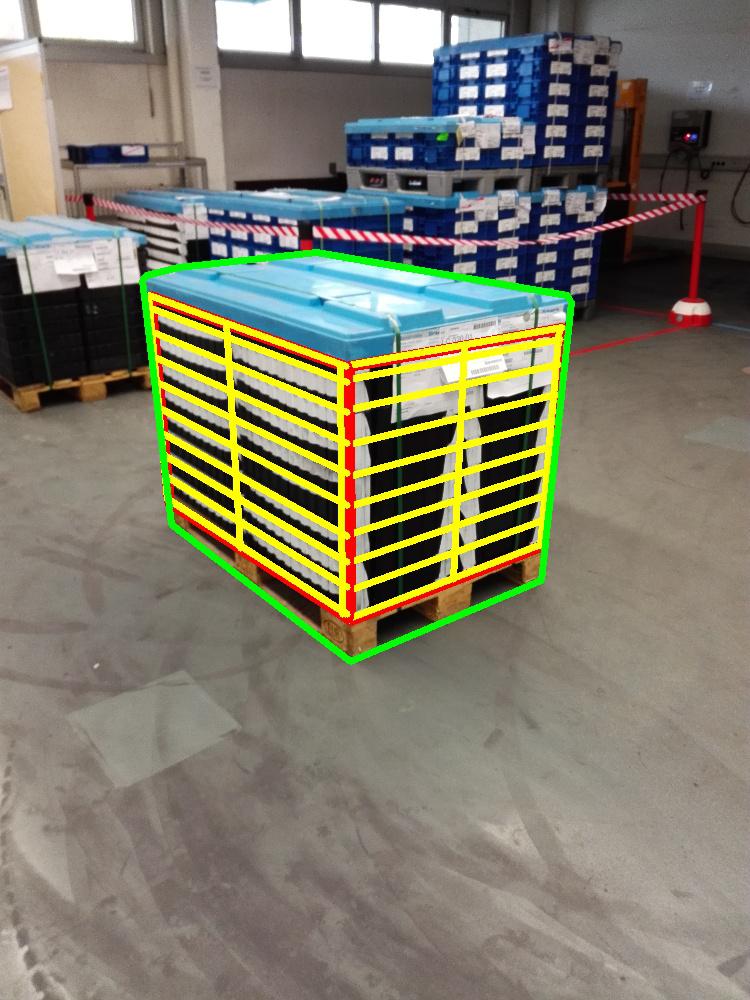}
	\includegraphics[width=0.19\linewidth,trim={0 2cm 0 2cm},clip=true]{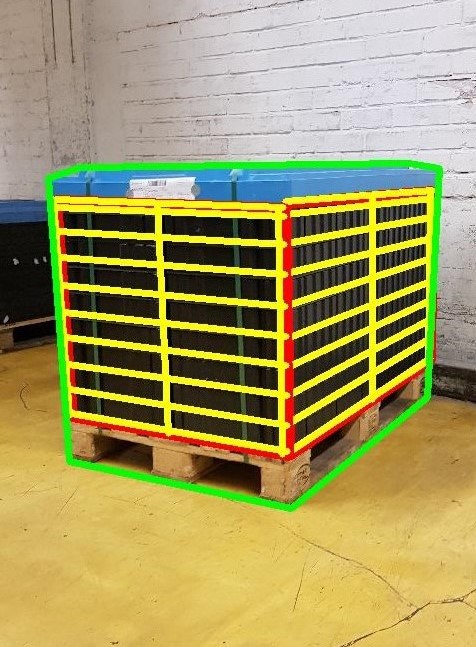}
	\caption{Evaluation image examples and result visualizations. (All of the images shown here were recognized correctly.)}
	\label{fig:evaluation_images}
\end{figure*}

\subsection{Training and Evaluation Dataset}
A dataset consisting of 1267 images of one or multiple transport units in logistics settings was acquired and annotated. 
The images were taken in the incoming goods department of a German component supplier in the automotive sector.
Data annotations consist of the following nested information:
\begin{itemize}
	\item Transport unit regions
	\item Transport unit side regions
	\item Packaging unit regions and class
	\item Base pallet regions and class
\end{itemize}
Each image contains up to three fully visible and annotated transport units, which are expected to be recognized. 
Additionally, arbitrary other transport units may be only partly visible within an image (these are not annotated).
Of the images available, 163 images containing 175 transport units are held back for evaluation of the packaging structure pipeline. 
The transport units contained in these images either consisted of KLT packages (112 images with one or two relevant transport units) or tray packages (51 images with one relevant transport unit each).
The other 1104 images were used for training and validation of the segmentation models.

\subsection{Segmentation Model Training and Evaluation}
All three segmentation models in use (transport unit segmentation, side and package segmentation, base pallet segmentation) are standard Mask-RCNN \cite{He2017:MaskRCNN} models using Inception-v2 \cite{Ioffe15:BatchNorm} feature extractors, as implemented in tensorflow's object detection API.
As our dataset of 1267 images is rather small and such approaches have proven to be a powerful solution in such cases, a transfer learning approach is used.
The models were initially trained on a the COCO object detection challenge's dataset \cite{Coco2018} of approximately 100 times the size of our data (123,287 annotated images).
Subsequently, the models were fine-tuned on 828 images for 100.000 training steps using gradient descent with momentum \cite{Qian1999:Momentum} and a batch-size of one.
Input image resolution was set to 600 pixels for the larger image dimension, transforming the image in such a way that original aspect ratio was preserved.
The Mask R-CNN's output mask resolution is set to 25 x 25 pixels.
Several image augmentation methods were used in neural network training, namely random horizontal flip, conversion to gray values, hue and brightness adjustments.
We experimented with additional augmentation methods, for instance random crop and pad, but found that they did not lead to improvements in accuracy.

For each model trained, the model's performance in accordance with the COCO Object Detection challenge's metric is measured; 
i.e. the mean Average Precision (mAP), averaged for ten different intersection over union (IoU) thresholds of 0.5 to 0.95, is computed.
Models were evaluated on the 163 dedicated evaluation images and on validation data, i.e. 25\% of the training image dataset that were not used in training.
The results are listed in Table \ref{tab:model_accuracies}:
For all three segmentation models the mAP is given. 
Additionally, the average precision for the individual classes (transport unit sides, KLT packaging units, tray packaging units) included in the side and package segmentation model is listed.

\begin{table}
	\caption{Segmentation Model mAP on Validation and Evaluation Data}
	\centering
	\begin{tabular}{|l|c|c|}
		\hline
		\textbf{Model / Class} & \textbf{Validation images} & \textbf{Evaluation images} \\
		\hline
		Transport Units & 0.954 & 0.979 \\ \hline
		Sides and Packages & 0.758 & 0.775 \\
		\ Sides & 0.877 & 0.892 \\
		\ KLT units & 0.740 & 0.764 \\
		\ Tray units & 0.659 & 0.668 \\ \hline
		Base pallets & 0.871 & 0.916 \\
		\hline
	\end{tabular}
	\label{tab:model_accuracies}
\end{table}

\subsection{Pipeline Evaluation}
The whole packaging structure process was evaluated on dedicated evaluation images, which were not used in model training or testing.
Exemplary evaluation images and result visualizations are shown in Fig. \ref{fig:evaluation_images}.

\subsubsection{Inter-Unit Segmentation}
In a first evaluation, we examined the results of the inter-unit segmentation as basis for the whole recognition pipeline by computing precision and recall of transport unit extractions.
A matching between groundtruth and detections was performed based on the masks' intersection over union, applying a threshold of 0.5.
The results are listed in Table \ref{tab:hu_errors}.
All 175 transport units were found correctly and only one false positive detection occurred.

\begin{table}
	\centering
	\caption{Evaluation Results: Inter-Unit Segmentation}
	\begin{tabular}{|l|c|c|}
		\hline
		Image Set & Precision & Recall \\
		\hline
		Evaluation All & 0.9943 & 1.0 \\
		KLT Images & 0.9920 & 1.0 \\
		Tray Images & 1.0 & 1.0 \\
		\hline
	\end{tabular}
	\label{tab:hu_errors}
\end{table}

\subsubsection{Recognition Pipeline}
We evaluate the whole packaging structure recognition pipeline by examining how many of the recognized transport units were recognized accurately.
At first, a recognition error $e_i$ for each evaluation image $i$ is computed independently as:
\begin{equation}
	e_i = 1 - \frac{\left|TP_i\right|}{\left|TP_i\right| + \left|FP_i\right| + \left|FN_i\right|}
\end{equation}
where $TP_i$, $FP_i$ and $FN_i$ are the sets of true positive, false positive and false negative results within the image.
In this context, true positive means, an annotated transport unit was found and the packaging structure was recognized correctly.
False positive can mean two things; either an additional transport unit was found where there was no unit annotated, or the packaging structure of an annotated transport unit was not recognized correctly.
As usual, false negatives are annotated transport unit's which were not recognized at all.
The overall evaluation error $e$ is computed as the mean error over all evaluation images:
\begin{equation}
	e = \sum_{i \in I} \frac{e_i}{|I|}
\end{equation}
where $I = \{1,...,163\}$ is the set of evaluation images.

\begin{table}
	\centering
	\caption{Evaluation Results: Mean Image Errors}
	\begin{tabular}{|l|l|c|}
		\hline
		Image Set & \# & Evaluation Error \\
		\hline
		Evaluation All & 163 & 0.1564 \\
		KLT Images & 112 & 0.0938 \\
		Tray Images & 51 & 0.2941 \\
		\hline
	\end{tabular}
	\label{tab:eval_errors}
\end{table}

An overview of the evaluation results is given in Table \ref{tab:eval_errors}.
The dataset's overall evaluation error is computed as 0.1564 indicating that approximately 85\% of all transport units were recognized correctly.
Accuracy is significantly higher for transport units with KLT packages compared to tray packages. 
As it is the case for the human eye, KLT units seem to be easier to recognize and differentiate for the neural network, as can also be seen in the result presented in Table \ref{tab:model_accuracies}.

\subsubsection{Manual Error Observations}
Apart from the quantitative evaluations presented above, insights and model understanding can be gained by qualitative observations.
To this end, some error cases are analyzed in this subsection.

\begin{figure}[htb]
	\centering
	\includegraphics[width=0.3\linewidth,trim={0 2cm 0 4cm},clip=true]{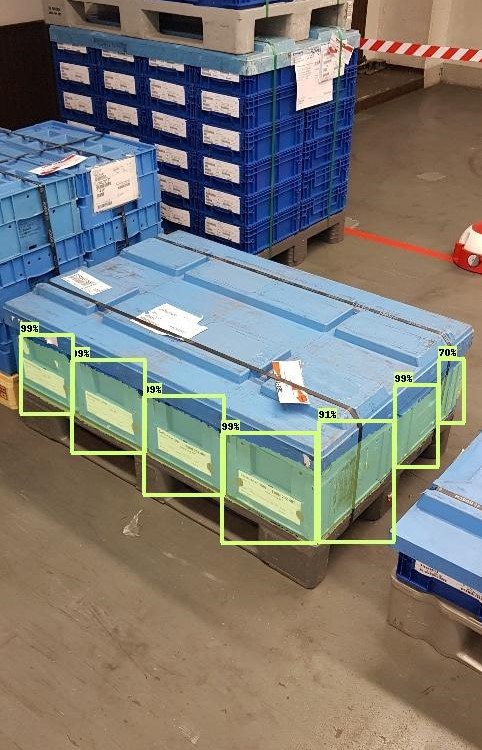}
	\includegraphics[width=0.3\linewidth,trim={0 2cm 0 4cm},clip=true]{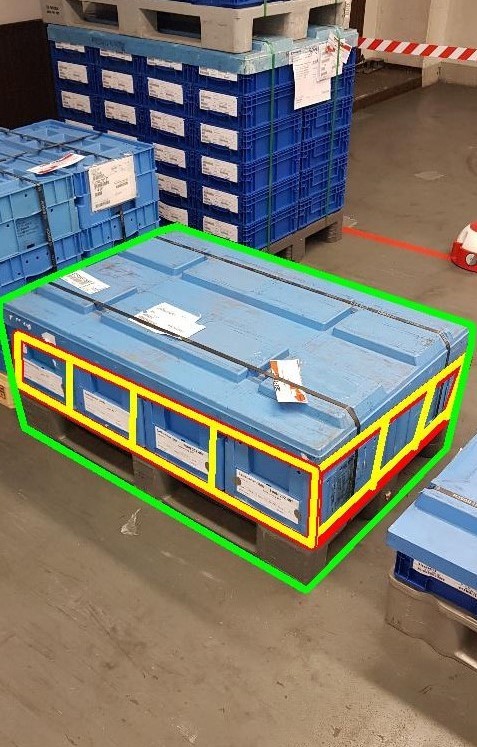}\\ \vspace{0.1cm}
	\includegraphics[width=0.3\linewidth,trim={0 3cm 0 5cm},clip=true]{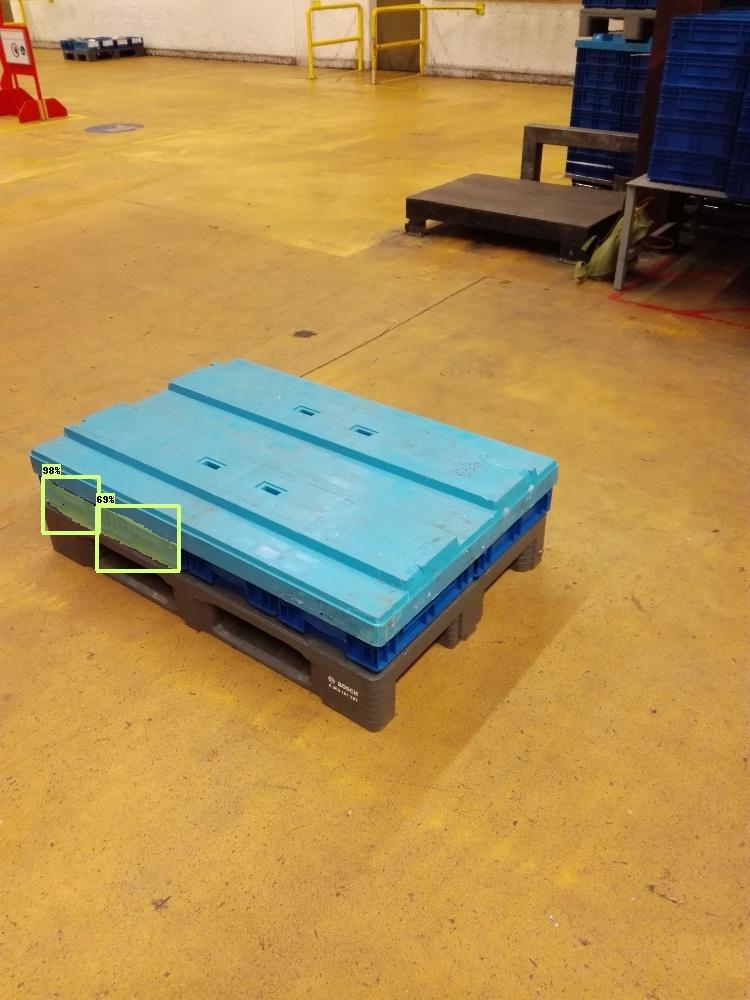}
	\includegraphics[width=0.3\linewidth,trim={0 3cm 0 5cm},clip=true]{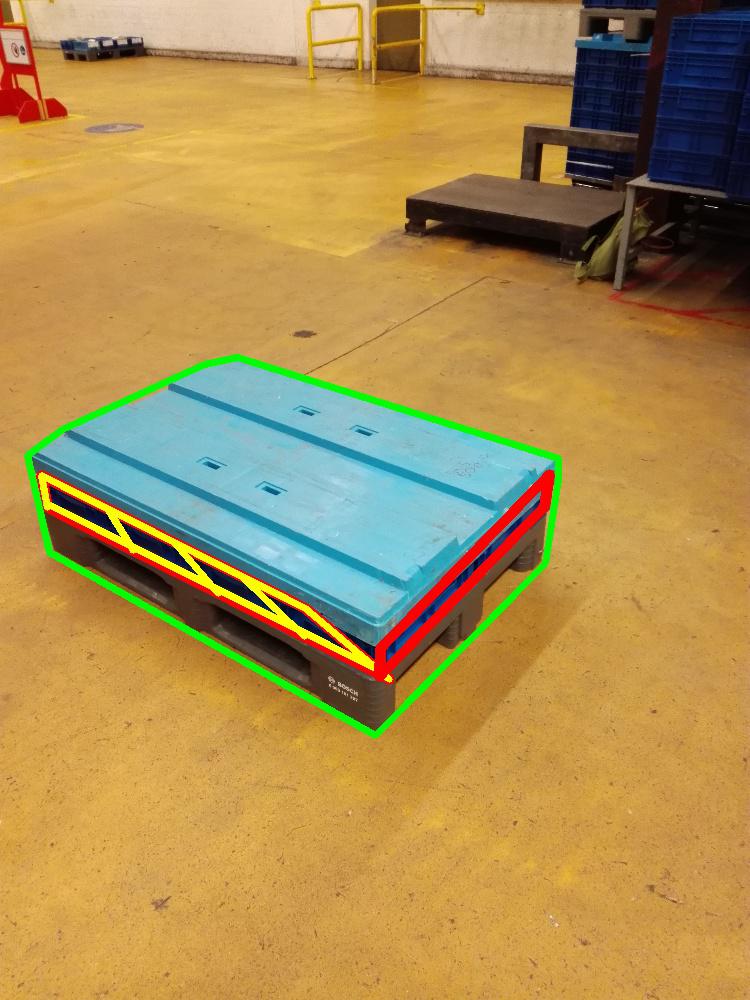}
	\caption{Error examples: single layer KLT transport units. Left: packaging unit detections. Right: Overall result visualization.}
	\label{fig:eval_error_observations}
\end{figure}

Altogether, 11 of 124 KLT transport units were not recognized correctly. 
Nine of these 11 errors were transport units consisting of one vertical layer of packaging units only, examples are shown in Fig. \ref{fig:eval_error_observations}.
The relatively high number of errors for such transport units can be explained by several reasons:
On one hand, the partly occluded transport units are harder to recognize for the segmentation model, especially in the case of less tall transport units where greater parts of the unit are occluded by the pallet lid.
The same holds for transport unit sides, which are smaller within the image and therefore harder to segment accurately.
Additionally, missing or flawed transport unit or side detections quickly lead to faulty results as the total package number is smaller. 
Recall here that package numbers are computed as the average package unit size relative to transport unit side size (see section \ref{sec:algorithm}).
Another reason are disadvantageous perspectives: 
As the total transport unit height of single layer units is low, images taken from a standing or kneeling position tend to show the transport unit from a top-down, rather than a frontal perspective.

\begin{figure}[htb]
	\centering
	\includegraphics[width=0.4\linewidth,trim={0 0 0 9cm},clip=true]{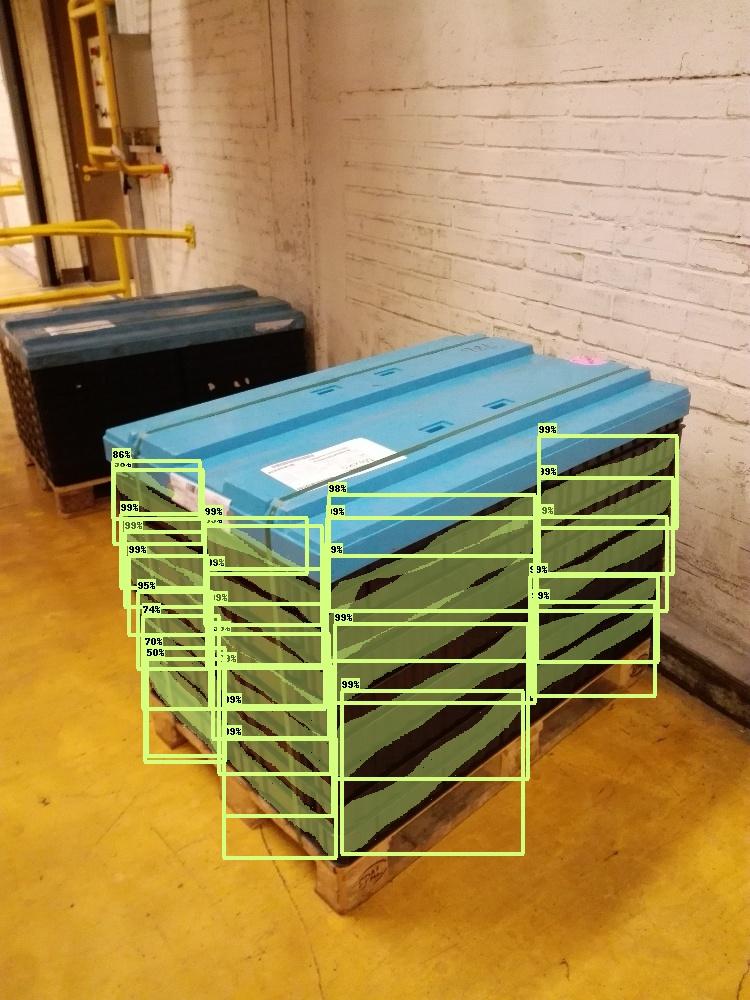}
	\includegraphics[width=0.4\linewidth,trim={0 0 0 9cm},clip=true]{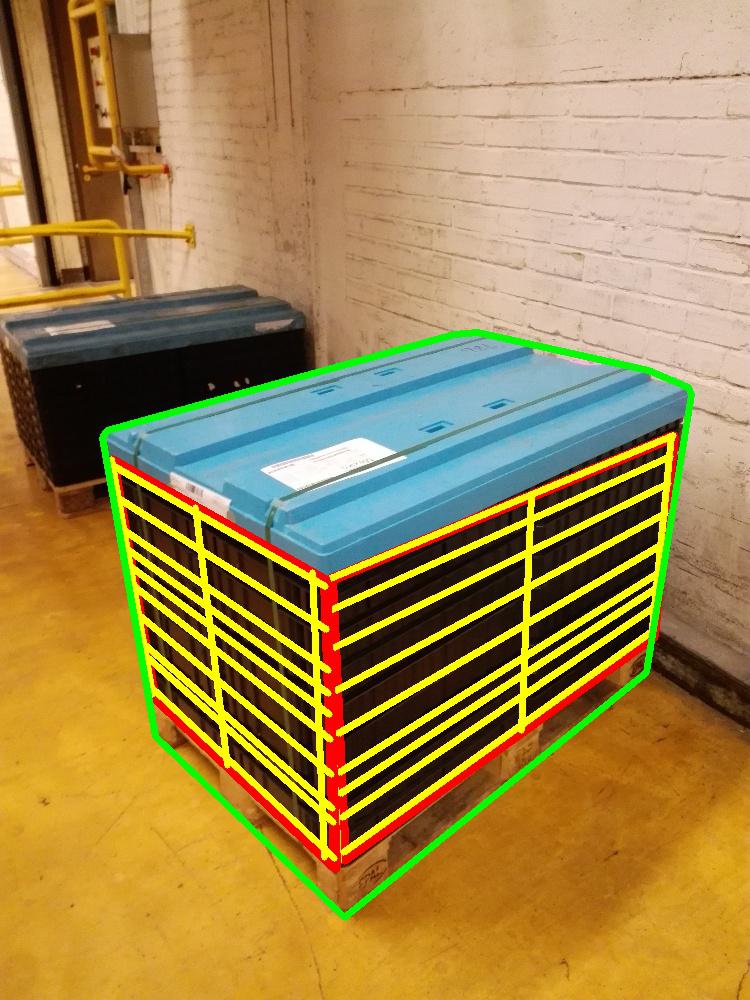}
	\caption{Error examples: Tray transport units. Left: packaging unit detections. Right: Overall result visualization.}
	\label{fig:eval_error_tray}
\end{figure}

The error rate for images showing transport units composed of tray packages was significantly higher than for KLT packages.
An example for faulty results is shown in Fig. \ref{fig:eval_error_tray}.
One reason is that it is substantially more difficult to visually segment the individual tray packaging units in images.
This is presumably not only due to the black color being disadvantageous, but even more to their structured surfaces and architecture.
In the case of tray packaging units, we even suspect that other approaches towards packaging unit segmentation, e.g. counting units based on a pixel value based edge detection, 
might be advantageous over neural-network-based instance segmentation.

\section{Summary and Outlook}
\label{sec:summary}

We presented a novel image processing pipeline for fully-automated packaging structure recognition applicable to various use-cases in logistics environments.
We trained our method on a small dataset to recognize a defined subset of logistics units, as relevant in supply chains of the automotive industry, for instance.
Evaluations show that the majority of transport units in the relevant setting can already be recognized accurately on single images, 
with the success rate of transport units with KLT packaging units being remarkably higher than those of tray packaging units.

One branch of our future work will be directed towards showing the applicability of our method to a broader range of logistics settings:
Based on additional annotated data, and also employing synthetically generated or augmented data, we aim to show that accurate packaging structure recognition is also possible for different and larger sets of packaging units and transport unit components.
On the other hand, various improvements and extensions to our method will be considered and evaluated.
We aim to train our method to recognize additional packaging components like pallet lids, packaging straps and transport labels.
Combining the information of multiple images of the same transport unit can lead to further accuracy improvements and allow for additional recognition of not completely uniformly, but regularly, packed transport units.
Further, we plan to integrate a unit measurement estimation employing given size information, such as sizes of base pallets, transport labels or background markers.
Methodically, we target the integration of a priori knowledge about transport unit composure (exploitation of rectangular and cubic shapes) to achieve further accuracy improvements.

\bibliography{bib}

\begin{thebibliography}{10}

\bibitem{akccay2016:transferAirportSecurity}
S.~Ak{\c{c}}ay, M.~E. Kundegorski, M.~Devereux, and T.~P. Breckon.
\newblock Transfer learning using convolutional neural networks for object
  classification within x-ray baggage security imagery.
\newblock In {\em 2016 IEEE International Conference on Image Processing
  (ICIP)}, pages 1057--1061. IEEE, 2016.

\bibitem{azizpour2015:transferlearning}
H.~Azizpour, A.~Sharif~Razavian, J.~Sullivan, A.~Maki, and S.~Carlsson.
\newblock From generic to specific deep representations for visual recognition.
\newblock In {\em Proceedings of the IEEE conference on computer vision and
  pattern recognition workshops}, pages 36--45, 2015.

\bibitem{Borstell18:ImageProcessingLogistics}
H.~Borstell.
\newblock A short survey of image processing in logistics - how image
  processing contributes to efficiency of logistics processes through
  intelligence.
\newblock {\em 11th International Doctoral Student Workshop on Logistics},
  2018.

\bibitem{Solution:cognex}
Cognex.
\newblock Logistics {Industry} {Solutions}.
\newblock \url{https://www.cognex.com/industries/logistics}.
\newblock Accessed: 2020-03-06.

\bibitem{Douglas1973:AlgorithmsFT}
D.~H. Douglas and T.~K. Peucker.
\newblock Algorithms for the reduction of the number of points required to
  represent a digitized line or its caricature.
\newblock {\em The Canadian Cartographer}, 10(2):112--122, 1973.

\bibitem{EPAL:pallet}
{European Pallet Association e.V. (EPAL)}.
\newblock {EPAL Euro Pallet (EPAL 1)}.
\newblock
  \url{https://www.epal-pallets.org/eu-en/load-carriers/epal-euro-pallet/}.
\newblock Accessed: 2020-02-26.

\bibitem{Furmans2019:Wareneingang}
K.~Furmans and C.~Kilger.
\newblock {\em Betrieb von Logistiksystemen}, pages 166--175.
\newblock Springer, 2019.

\bibitem{gao2018:transferEnviornment}
Y.~Gao and K.~M. Mosalam.
\newblock Deep transfer learning for image-based structural damage recognition.
\newblock {\em Computer-Aided Civil and Infrastructure Engineering},
  33(9):748--768, 2018.

\bibitem{VDA:KLT}
{German Association of the Automotive Industry}.
\newblock {VDA 4500 - Small Load Carrier (SLC) System (KLT)}.
\newblock
  \url{https://www.vda.de/en/services/Publications/small-load-carrier-(slc)-system-(klt).html}.
\newblock Accessed: 2020-01-29.

\bibitem{He2017:MaskRCNN}
K.~He, G.~Gkioxari, P.~Doll{\'{a}}r, and R.~B. Girshick.
\newblock Mask {R-CNN}.
\newblock {\em CoRR}, abs/1703.06870, 2017.

\bibitem{BVL:Digitalisierung}
F.~Heistermann, M.~ten Hompel, and T.~Mallée.
\newblock Bvl positionspapier: Digitalisierung in der logistik.
\newblock \url{https://www.bvl.de/positionspapier-digitalisierung}, 2017.

\bibitem{Solution:Fraunhofer_Ladungstraeger}
J.~Hinxlage and J.~Möller.
\newblock Ladungsträgerzahlung per smartphone.
\newblock {\em Jahresbericht Fraunhofer IML 2018}, pages 72--73, 2018.

\bibitem{huh2016:transferlearningImagenet}
M.~Huh, P.~Agrawal, and A.~A. Efros.
\newblock What makes imagenet good for transfer learning?
\newblock {\em arXiv preprint arXiv:1608.08614}, 2016.

\bibitem{Ioffe15:BatchNorm}
S.~Ioffe and C.~Szegedy.
\newblock Batch normalization: Accelerating deep network training by reducing
  internal covariate shift.
\newblock {\em CoRR}, abs/1502.03167, 2015.

\bibitem{Coco2018}
T.~Lin, M.~Maire, S.~J. Belongie, L.~D. Bourdev, R.~B. Girshick, J.~Hays,
  P.~Perona, D.~Ramanan, P.~Doll{\'{a}}r, and C.~L. Zitnick.
\newblock Microsoft {COCO:} common objects in context.
\newblock {\em CoRR}, abs/1405.0312, 2014.

\bibitem{Solution:Logivations}
Logivations.
\newblock Ki-basierte identifikation in der logistik.
\newblock
  \url{https://www.logivations.com/de/solutions/agv/camera_identification.php#count_and_measure}.
\newblock Accessed: 2020-03-06.

\bibitem{Minaee2020:ImageSegmentationSurvey}
S.~Minaee, Y.~Boykov, F.~M. Porikli, A.~J. Plaza, N.~Kehtarnavaz, and
  D.~Terzopoulos.
\newblock Image segmentation using deep learning: A survey.
\newblock {\em ArXiv}, abs/2001.05566, 2020.

\bibitem{pan2009:transferlearningsurvey}
S.~J. Pan and Q.~Yang.
\newblock A survey on transfer learning.
\newblock {\em IEEE Transactions on knowledge and data engineering},
  22(10):1345--1359, 2009.

\bibitem{Qian1999:Momentum}
N.~Qian.
\newblock On the momentum term in gradient descent learning algorithms.
\newblock {\em Neural Netw.}, 12(1):145–151, Jan. 1999.

\bibitem{russakovsky2015:imagenet}
O.~Russakovsky, J.~Deng, H.~Su, J.~Krause, S.~Satheesh, S.~Ma, Z.~Huang,
  A.~Karpathy, A.~Khosla, M.~Bernstein, et~al.
\newblock Imagenet large scale visual recognition challenge.
\newblock {\em International journal of computer vision}, 115(3):211--252,
  2015.

\bibitem{shin2016:transferMedicalImaging}
H.-C. Shin, H.~R. Roth, M.~Gao, L.~Lu, Z.~Xu, I.~Nogues, J.~Yao, D.~Mollura,
  and R.~M. Summers.
\newblock Deep convolutional neural networks for computer-aided detection: Cnn
  architectures, dataset characteristics and transfer learning.
\newblock {\em IEEE transactions on medical imaging}, 35(5):1285--1298, 2016.

\bibitem{Solution:vitronic}
Vitronic.
\newblock Warehouse logistics \& distribution logistics.
\newblock
  \url{https://www.vitronic.com/industrial-and-logistics-automation/sectors/warehouse-logistics-distribution-logistics.html}.
\newblock Accessed: 2020-03-06.

\bibitem{Wang2019:fewshotlearning}
Y.~Wang and Q.~Yao.
\newblock Few-shot learning: {A} survey.
\newblock {\em CoRR}, abs/1904.05046, 2019.

\bibitem{Wei19:DigitalTechnologiesLogistics}
F.~Wei, C.~Alias, and B.~Noche.
\newblock {\em Applications of Digital Technologies in Sustainable Logistics
  and Supply Chain Management}, pages 235--263.
\newblock Innovative Logistics Services and Sustainable Lifestyles:
  Interdependencies, Transformation Strategies and Decision Making. Springer
  International Publishing, Cham, 2019.

\bibitem{Solution:zetes}
Zetes.
\newblock Zetes: {Warenannahme} und {Versand}.
\newblock \url{https://www.zetes.com/de/technologien-supplies/machine-vision}.
\newblock Accessed: 2020-03-06.

\end{thebibliography}

\end{document}